\documentclass{article}
\usepackage{EMNLP2023}
\usepackage{times}
\usepackage{latexsym}
\usepackage[T1]{fontenc}
\usepackage[utf8]{inputenc}
\usepackage{microtype}
\usepackage{graphicx}
\usepackage{multirow}
\usepackage{bm}
\usepackage{multirow}
\usepackage{caption}
\usepackage{natbib}
\usepackage{amsfonts}
\usepackage{amsmath}
\usepackage{subfigure}
\usepackage{booktabs}
\usepackage{times}
\usepackage{latexsym}
\usepackage{hyperref}
\usepackage{tabularx}
\usepackage{adjustbox}
\usepackage{tipa} 
\usepackage{enumitem}
\usepackage{amssymb}
\usepackage{amsthm}
\usepackage{pifont} 
\usepackage{array}
\usepackage{comment}
\usepackage{amsmath}
\usepackage{mathtools}

\usepackage[T4]{fontenc}
\usepackage{newunicodechar}
\newenvironment{tfour}{\fontencoding{T4}\selectfont}{}

\newunicodechar{ɔ}{\fon}
\newunicodechar{ɖ}{\fon}

\usepackage[T1]{fontenc}
\usepackage[utf8]{inputenc}
\usepackage{microtype}

\title{FonMTL: Towards Multitask Learning for the Fon Language}

\author{\normalsize Bonaventure F. P. Dossou$^{1,2,3,4*}$, Iffanice Houndayi$^{5*}$, Pamely Zantou$^{5}$, Gilles Hacheme$^{6*}$\\ 
\\\\
\footnotesize
$^*$Masakhane NLP, $^1$Mila Quebec AI Institute, $^2$ McGill University, $^3$ Lelapa AI, $^4$ Lanfrica\\
\footnotesize
$^5$Carnegie Mellon University Africa,
$^6$Microsoft AI For Good Kenya}

\begin{document}
\maketitle
\begin{abstract}
The Fon language, spoken by an average 2 million of people, is a truly low-resourced African language, with a limited online presence, and existing datasets (just to name but a few). Multitask learning is a learning paradigm that aims to improve the generalization capacity of a model by sharing knowledge across different but related tasks: this could be prevalent in very data-scarce scenarios. In this paper, we present the first explorative approach to multitask learning, for model capabilities enhancement in Natural Language Processing for the Fon language. Specifically, we explore the tasks of Named Entity Recognition (NER) and Part of Speech Tagging (POS) for Fon. We leverage two language model heads as encoders to build shared representations for the inputs, and we use linear layers blocks for classification relative to each task. Our results on the NER and POS tasks for Fon, show competitive (or better) performances compared to several multilingual pretrained language models finetuned on single tasks. Additionally, we perform a few ablation studies to leverage the efficiency of two different loss combination strategies and find out that the equal loss weighting approach works best in our case. Our code is open-sourced at \url{https://github.com/bonaventuredossou/multitask_fon}.
\end{abstract}

\section{Introduction}
Learning one task at a time might be ineffective and prone to overfitting, low data efficiency, or slow learning \citep{zhang2022survey}, especially for large and complex problems. This might sometimes lead to the training of very dedicated models with generalization capacity only local to the tasks they have been trained on. Multitask Learning (MTL), precisely for that reason targets a more efficient way of learning, by sharing common representations of the inputs and by implicitly transferring information among the various tasks \citep{caruana1993multitask, Caruana_1997}. MTL efficiency and promises have been demonstrated in the literature \citep{gessler-zeldes-2022-microbert, ruder2017overview} along with the benefit of the understanding of tasks (their similarity, relationship, hierarchy) for MTL.

For African low-resourced languages in general, building NLP models on specific tasks can be in general difficult for several reasons, including the availability of datasets \citep{joshi-etal-2020-state, emezue-dossou-2021-mmtafrica, nekoto-etal-2020-participatory, ffr}. This is even more true for the Fon language (see Appendix \ref{sec:fon_language}), a truly low-resourced African language. Therefore, MTL appears to be a promising approach to exploring downstream tasks in the Fon language.
\raggedbottom

The task of Part-of-Speech (POS) tagging is to assign grammatical groups (i.e., whether it is a noun, an adjective, a verb, an adverb, and more) to words in a sentence using contextual cues and assigning corresponding tags. On the other hand, Named Entity Recognition (NER) tries to find out whether or not a word is a named entity (persons, locations, organizations, time expressions, and more). This problem is twofold: detection of names and categorizing names. Consequently, whereas POS is more of a global problem since there can be relationships between the first and the last word of a sentence, NER is rather local, as named entities are not spread in a sentence and mostly consist of uni, bi, or trigrams. Albeit different, and adding the fact that both tasks are classification tasks, we then speculate that both tasks could benefit each other.
\section{Experiments and Results}
In this paper, we explore the first multitask learning model for the Fon language, particularly, focusing on the NER and POS tasks. We used a hard parameter sharing approach --- the most popular MTL method used in the literature \citep{caruana1993multitask, ruder2017overview} and falls into the joint training method described by \citep{zhang2022survey} in their taxonomy of MTL approaches for NLP. 

\paragraph{Experiments:} For NER, we used the Fon dataset of MaskhaNER 2.0 \citep{adelani-etal-2022-masakhaner}, a NER dataset of 20 African languages, including Fon. For the POS task, we used the MasakhaPOS dataset \citep{dione2023masakhapos}, the largest POS dataset for 20 typologically diverse African languages, including Fon. In MassakhaNER 2.0, we have 4343/621/1240 as train/dev/test set sizes, with a total of 173099 tokens. In MasakhaPOS we have 798/159/637 as train/dev/test set sizes, with a total of 49460 tokens and 30.6 tokens on average per sentence. For our experiments we leveraged two setups:
\begin{itemize}
    \item pre-train on all languages of both datasets, and evaluate on the Fon test sets (for NER and POS),
    \item pre-train only on Fon data from both datasets, and evaluate on the Fon test sets.
\end{itemize}

\begin{figure}[!ht]
  \centering
  \includegraphics[width=0.35\textwidth]{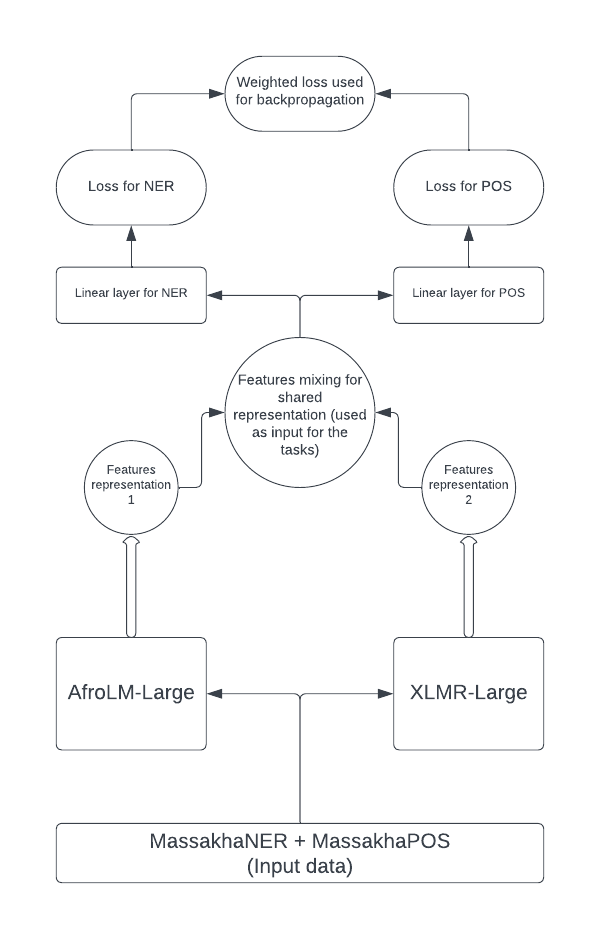}
  \caption{Our model architecture}
  \label{fig:modelarch}
\end{figure}

In our MTL model (Figure \ref{fig:modelarch}), we use two language model (LM) heads: AfroLM-Large \citep{dossou-etal-2022-afrolm} and XLMR-Large \citep{conneau-etal-2020-unsupervised}. AfroLM has been pretrained from scratch on 23 African languages in an active learning framework (see Section 3 in \citep{dossou-etal-2022-afrolm, dossou2023adapting}), while XLMR-Large has been pretrained on 100 languages (with $\leq$5 African). Each LM head is used to build representations of the inputs from each task, and both representations are then combined (in a multiplicative way) to build a shared representation across models and tasks. Following the shared representation, we built two linear layers, serving as classification layers for each respective task.
\raggedbottom

We combined the cross-entropy losses of both tasks in two ways: (a) unweighted sum \citep{NEURIPS2022_4f301ae9} i.e., the unitary sum of each individual loss, and (b) equally weighted where losses are weighted by two parameters $\alpha$ and $\beta$ then added up together. For simplicity, in our experiments, we set $\alpha=\beta=0.5$. We used a small batch size of 4, the AdamW \citep{loshchilov2017decoupled} as optimizer with a learning rate of 3e-5.

\paragraph{Results:} Along with our MTL variants, we evaluated several multilingual pretrained language models (MPLMs) on the NER task: AfroXLMR \citep{alabi-etal-2022-adapting} (base and large), AfroLM-Large \citep{dossou-etal-2022-afrolm}, AfriBerta-Large \citep{ogueji-etal-2021-small}, and XLMR \citep{conneau-etal-2020-unsupervised} (base and large). AfriBerta is a multilingual model pretrained on 11 African languages, while AfroXLMR is an adapted version of XLMR to 17 African Languages via Multilingual Adaptive Fine-Tuning.

In Tables \ref{tab:results} and \ref{tab:pos_results}, we can see that our MTL models are competitive, and in some cases outperformed some single-task baselines; for both NER and POS tasks: This means that as we speculated earlier, both tasks data and representations are informative and useful for the downstream performances. Our results are also backed-up by several empirical results \citep{liu2019multi, crawshaw2020multi, dwa} on various tasks like Natural Language Understanding (NLU) and GLUE tasks. Furthermore, we can see that the equal-weighted sum of both losses worked better than the unweighted sum \citep{dwa, dynamic_loss}, which makes sense as we treated both tasks \textit{equally}. It is also important to note that the efficiency of our model (on both tasks), when trained on all languages is higher than when trained solely on the data for Fon: this means that Fon benefited from the information from other languages. This could be due to various factors (features) like geographical distance, phonological distance, entity overlap, just to name but a few \cite{adelani-etal-2022-masakhaner}.
\begin{table}
\vspace{-1mm}
\centering
\caption{\textbf{NER Performance}: F1-Score of several multilingual pretrained language models (baselines \& single task) versus our MTL variants, on the Fon NER test set (after 50 epochs). MTL Sum is the MTL model with the unweighted sum of both losses, and MTL Weighted is the model with an equally weighted sum of both losses. $^{*}$ represents the results when the data from all languages of the MasakhaNER dataset has been used for training or finetuning; while $^{+}$ represents the results when the only data of the Fon language has been used for training.}
\resizebox{0.5\columnwidth}{!}{
\begin{tabular}{ll}
\toprule
\textbf{Models} & \textbf{F1-Score} \\ \midrule
AfroLM-Large (single task; baseline; $^{*}$) & 80.48       \\ 
AfriBerta-Large (single task; baseline; $^{*}$) & 79.90      \\ 
XLMR-Base (single task; baseline; $^{*}$) & 81.90                \\ 
XLMR-Large (single task; baseline; $^{*}$) & 81.60                \\
AfroXLMR-Base (single task; baseline; $^{*}$) & 82.30                \\ 
AfroXLMR-Large (single task; baseline; $^{*}$) & 82.70                \\ \midrule 
MTL Sum (multi-task; ours; $^{*}$) & \textcolor{red}{\textbf{79.87}}               \\ 
MTL Weighted (multi-task; ours; $^{*}$) & \textcolor{red}{\textbf{81.92}}
       \\ \midrule 
MTL Weighted (multi-task; ours; $^{+}$) & \textcolor{red}{\textbf{64.43}}       \\ \bottomrule
\end{tabular}
}
\label{tab:results}
\vspace{-4mm}
\end{table}

\begin{table}
\vspace{3mm}
\centering
\caption{\textbf{POS Performance}: Accuracy Score of several multilingual pretrained language models (baselines \& single task) versus our MTL variants, on the Fon POS test set (after 50 epochs).}
\resizebox{0.5\columnwidth}{!}{
\begin{tabular}{ll}
\toprule
\textbf{Models} & \textbf{Accuracy Score} \\ \midrule
AfroLM-Large (single task; baseline; $^{*}$) &  82.40       \\ 
AfriBerta-Large (single task; baseline; $^{*}$) & 88.40      \\ 
XLMR-Base (single task; baseline; $^{*}$) & 90.10                \\ 
XLMR-Large (single task; baseline; $^{*}$) & 90.20                \\
AfroXLMR-Base (single task; baseline; $^{*}$) & 90.10                \\ 
AfroXLMR-Large (single task; baseline; $^{*}$) & 90.40                \\ \midrule 
MTL Sum (multi-task; ours; $^{*}$) & \textcolor{red}{\textbf{82.45}}               \\ 
MTL Weighted (multi-task; ours; $^{*}$) & \textcolor{red}{\textbf{89.20}}
       \\ \midrule 
MTL Weighted (multi-task; ours; $^{+}$) & \textcolor{red}{\textbf{80.85}}       \\ \bottomrule
\end{tabular}
}
\label{tab:pos_results}
\vspace{-4mm}
\end{table}

\begin{table}
\vspace{3mm}
\centering
\caption{\textbf{Merging Representation Type Performance}: Impact of representation merging type, on the downstream task performance}
\resizebox{\columnwidth}{!}{
\begin{tabular}{lllll}
\toprule
\textbf{Merging Type} & \textbf{Models} & \textbf{Task} & \textbf{Metric} & \textbf{Metric Value} \\ \midrule
Multiplicative & MTL Weighted (multi-task; ours; $^{*}$) & NER & F1-Score & {\textbf{81.92}} \\
Multiplicative & MTL Weighted (multi-task; ours; $^{+}$) & NER & F1-Score & 64.43 \\ \midrule
Multiplicative & MTL Weighted (multi-task; ours; $^{*}$) & POS & Accuracy & {\textbf{89.20}} \\
Multiplicative & MTL Weighted (multi-task; ours; $^{+}$) & POS & Accuracy & 80.85   \\ \bottomrule

Additive & MTL Weighted (multi-task; ours; $^{*}$) & NER & F1-Score & 78.91 \\
Additive & MTL Weighted (multi-task; ours; $^{+}$) & NER & F1-Score & 60.93 \\ \midrule
Additive & MTL Weighted (multi-task; ours; $^{*}$) & POS & Accuracy & 86.99 \\
Additive & MTL Weighted (multi-task; ours; $^{+}$) & POS & Accuracy & 78.25 \\ \bottomrule
\end{tabular}
}
\label{tab:merging_results}
\vspace{-4mm}
\end{table}
Our last analysis consisted of comparing the most efficient way of merging the representations from both language model heads. The results are presented in Table \ref{tab:merging_results}. We can see that the \textbf{multiplicative} approach provided better results than the \textit{additive} one. Our intuition is that multiplication is closer to the scale-dot product (which is a type of attention mechanism), which captures well vectorial relationships; thus providing a meaningful unified representation, that is useful on the downstream tasks.
\section{Conclusion}
In this paper, we presented the first effort to leverage MTL for NLP downstream tasks in Fon. Our results on Fon NER and POS tasks showed competitive (and sometimes better) performances of our MTL methods. We hope these results enable more exploration of MTL in NLP for Fon in particular, and African Languages in general. In future works, we want to explore: (a) the impact of merging representations in an additive way, and (b) the dynamic weighted average loss \citep{dynamic_loss, dwa} for our training loop; which has been empirically demonstrated to be effective. Additionally, as our results showed that training with the data from all languages, helped the downstream performances (on both tasks), we would like to explore why is this the case i.e., answering the question: ``given a target downstream task language $\mathcal{X}$, which other language(s) and their data, would be beneficial to $\mathcal{X}$? How and why``? Our code is open-sourced at \url{https://github.com/bonaventuredossou/multitask_fon}.

\bibliography{mtl}
\bibliographystyle{acl_natbib}

\appendix
\section{Linguistic Description of Fon}
\label{sec:fon_language}
Fon also known as Fongbé is a native language of the Benin Republic. It is spoken on average by 1.7 million people. Fon belongs to the \textit{Niger-Congo-Gbe} languages family. It is a tonal, isolating, and left-behind language according to \citep{joshi}, with an \textit{Subject-Verb-Object} (SVO) word order. Fon has about 53 different dialects, spoken throughout Benin \citep{grammaire, fon_phonology, ethnologue}. Its alphabet is based on the Latin alphabet, with the addition of the letters: \begin{tfour}\m{o}\end{tfour}, \begin{tfour}\m{d}\end{tfour}, \begin{tfour}\m{e}\end{tfour}, and the digraphs gb, hw, kp, ny, and xw. There are 10 vowels phonemes in Fon: 6 said to be closed [i, u, ĩ, ũ], and 4 said to be opened [\begin{tfour}\m{e}\end{tfour}, \begin{tfour}\m{o}\end{tfour}, a, ã]. There are 22 consonants (m, b, n, \begin{tfour}\m{d}\end{tfour}, p, t, d, c, j, k, g, kp, gb, f, v, s, z, x, h, xw, hw, w). Fon has two phonemic tones: high and low. High is realized as rising \textit{(low–high)} after a consonant. Basic disyllabic words have all four possibilities: \textit{high-high}, \textit{high-low}, \textit{low-high}, and \textit{low-low}. In longer phonological words, like verb and noun phrases, a high tone tends to persist until the final syllable. If that syllable has a phonemic low tone, it becomes falling \textit{(high–low)}. Low tones disappear between high tones, but their effect remains as a downstep. Rising tones \textit{(low–high)} simplify to high after high (without triggering downstep) and to low before high \citep{grammaire, fon_phonology}.
\end{document}